\documentstyle[colacl,epsf]{article}

\title{Word Clustering and Disambiguation Based on Co-occurrence Data}
\author{Hang Li \and Naoki Abe\\ Theory NEC Laboratory, 
Real World Computing Partnership \\
c/o C\&C Media Research Laboratories., NEC\\
 4-1-1 Miyazaki, Miyamae-ku, Kawasaki 216-8555, Japan\\
\{lihang,abe\}@ccm.cl.nec.co.jp}

\begin{document}

\maketitle

\begin{abstract} We address the problem of clustering words (or
constructing a thesaurus) based on co-occurrence data, and using the
acquired word classes to improve the accuracy of syntactic
disambiguation.  We view this problem as that of estimating a joint
probability distribution specifying the joint probabilities of word
pairs, such as noun verb pairs. We propose an efficient algorithm
based on the Minimum Description Length (MDL) principle for estimating
such a probability distribution. Our method is a natural extension of
those proposed in \cite{Brown92} and \cite{Li96b}, and overcomes their
drawbacks while retaining their advantages.  We then combined this
clustering method with the disambiguation method of \cite{Li95} to
derive a disambiguation method that makes use of both automatically
constructed thesauruses and a hand-made thesaurus. The overall
disambiguation accuracy achieved by our method is $85.2\%$, 
which compares favorably against 
the accuracy ($82.4\%$) obtained by the state-of-the-art
disambiguation method of \cite{Brill94}.  \end{abstract}

\section{Introduction} We address the problem of clustering words, or
that of constructing a thesaurus, based on co-occurrence data.  We
view this problem as that of estimating a joint probability
distribution over word pairs, specifying the joint probabilities of
word pairs, such as noun verb pairs. In this paper, we assume that the
joint distribution can be expressed in the following manner, which is
stated for noun verb pairs for the sake of readability: The joint
probability of a noun and a verb is expressed as the product of the
joint probability of the noun class and the verb class which the
noun and the verb respectively belong to, and the conditional
probabilities of the noun and the verb given their respective
classes.

As a method for estimating such a probability distribution, we propose
an algorithm based on the Minimum Description Length (MDL) principle.
Our clustering algorithm iteratively merges noun classes
and verb classes in turn, in a bottom up fashion. For each merge it
performs, it calculates the increase in data description length
resulting from merging any noun (or verb) class pair, and performs the
merge having the least increase in data description length, provided
that the increase in data description length is less than the
reduction in model description length.

There have been a number of methods proposed in the literature to
address the word clustering problem (e.g.,
\cite{Brown92,Pereira93,Li96b}). The method proposed in this paper is
a natural extension of both Li \& Abe's and Brown et al's methods, and
is an attempt to overcome their drawbacks while retaining their advantages.

The method of Brown et al, which is based on the Maximum Likelihood
Estimation (MLE), performs a merge which would result in the least
reduction in (average) mutual information.  Our method turns out to be
equivalent to performing the merge with the least reduction in mutual
information, provided that the reduction is below a certain threshold
which depends on the size of the co-occurrence data and the number of
classes in the current situation. This method, based on the MDL
principle, takes into account both the fit to data and the simplicity
of a model, and thus can help cope with the over-fitting problem that
the MLE-based method of Brown et al faces.

The model employed in \cite{Li96b} is based on the assumption that the
word distribution within a class is a uniform distribution, i.e. every
word in a same class is generated with an equal probability.
Employing such a model has the undesirable tendency of classifying
into different classes those words that have similar co-occurrence
patterns but have different absolute frequencies. The proposed method, in
contrast, employs a model in which different words within a same class
can have different conditional generation probabilities, and thus
can classify words in a way that is not affected by words' absolute
frequencies and resolve the problem faced by the method of \cite{Li96b}.

We evaluate our clustering method by using the word classes and the
joint probabilities obtained by it in syntactic disambiguation
experiments. Our experimental results indicate that using the word
classes constructed by our method gives better disambiguation results
than when using Li \& Abe or Brown et al's methods.  By combining
thesauruses automatically constructed by our method and an existing
hand-made thesaurus (WordNet), we were able to achieve the overall
accuracy of $85.2\%$ for pp-attachment disambiguation, which compares 
favorably against 
the accuracy ($82.4\%$) obtained using the state-of-the-art
method of \cite{Brill94}.

\section{Probability Model} 
\begin{figure}[htb] 
\begin{center}
\epsfxsize4cm\epsfysize4cm\epsfbox{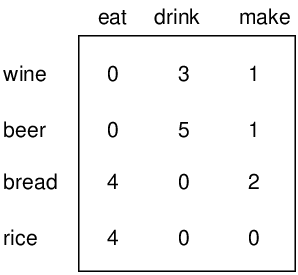}
\vspace{-5pt}
\caption{Example co-occurrence data} 
\label{fig:co-occurrence} 
\end{center}
\end{figure}

Suppose available to us are co-occurrence
data over two sets of words, such as the sample of verbs and the head
words of their direct objects given in
Fig.~\ref{fig:co-occurrence}. Our goal is to (hierarchically) cluster
the two sets of words so that words having similar co-occurrence
patterns are classified in the same class, and output a thesaurus for
each set of words.

We can view this problem as that of estimating the best probability model 
from among a class of models of (probability distributions) which can 
give rise to the co-occurrence data.

\begin{figure}[htb]
\begin{center}
\epsfxsize8cm\epsfysize7cm\epsfbox{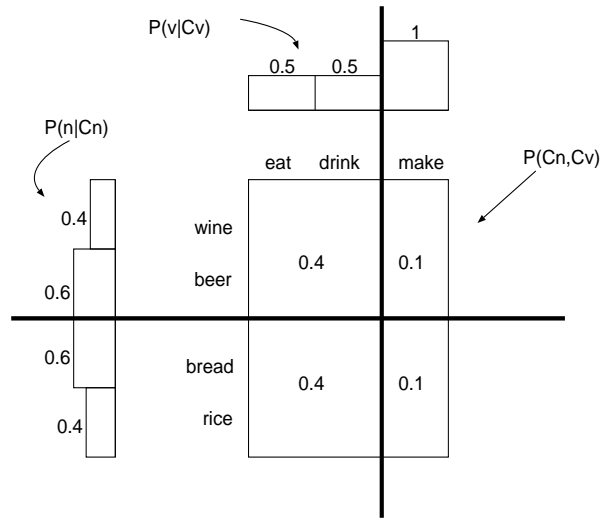}
\vspace{-5pt}
\caption{Example hard clustering model}
\label{fig:hardmodel} 
\end{center}
\end{figure}

In this paper, we consider the following type of probability
models. Assume without loss of generality that the two sets of words
are a set of nouns ${\cal N}$ and a set of verbs ${\cal V}$. A
partition $T_n$ of ${\cal N}$ is a set of noun-classes satisfying
$\cup_{C_n \in T_n} C_n = {\cal N}$ and $\forall C_i, C_j \in T_n,
C_i\cap C_j =\emptyset$. A partition $T_v$ of ${\cal V}$ can be
defined analogously. We then define a probability model of noun-verb
co-occurrence by defining the joint probability of a noun $n$ and a
verb $v$ as the product of the joint probability of the noun and verb
classes that $n$ and $v$ belong to, and the conditional probabilities
of $n$ and $v$ {\em given} their classes, that is,
\begin{equation}\label{eq:hardmodel} \begin{array}{c} P(n,v) =
P(C_n,C_v) \cdot P(n|C_n) \cdot P(v|C_v), \\ \end{array}
\end{equation} where $C_n$ and $C_v$ denote the (unique) classes to
which $n$ and $v$ belong. In this paper, we refer to this model as the
`hard clustering model,' since it is based on a type of clustering in
which each word can belong to only one class. Fig.~\ref{fig:hardmodel}
shows an example of the hard clustering model that can give rise to
the co-occurrence data in Fig.~\ref{fig:co-occurrence}.

\section{Parameter Estimation}\label{sec:estimate}
A particular choice of partitions for a hard clustering model 
is referred to as a `discrete' hard-clustering model, with 
the probability parameters left to be estimated.  
The values of these parameters can be estimated 
based on the co-occurrence data by the Maximum Likelihood Estimation. 
For a given set of co-occurrence data \[ {\cal
S} = \{(n_1,v_1),(n_2,v_2),\ldots,(n_m,v_m)\}, \] the maximum
likelihood estimates of the parameters are defined as the values that
maximize the following likelihood function with respect to the data:
\[
  \prod_{i=1}^{m} P(n_i,v_i) =  \prod_{i=1}^{m} 
  (P(n_i|C_{n_i}) \cdot P(v_i|C_{v_i}) \cdot P(C_{n_i},C_{v_i})).
\]

It is easy to see that this is possible by setting the parameters as 
\[
\hat{P}(C_n,C_v)= \frac{f(C_n,C_v)}{m}, \] 
\[ \forall x \in {\cal N} \cup {\cal V},  
\hat{P}(x|C_x)= \frac{f(x)}{f(C_x).} 
\]
Here, $m$ denotes the entire data size, $f(C_n,C_v)$ the frequency of
word pairs in class pair $(C_n,C_v)$, $f(x)$ the frequency of word
$x$, and $f(C_x)$ the frequency of words in class $C_x$.

\section{Model Selection Criterion}
The question now is what criterion should we employ to select the best
model from among the possible models.  Here we adopt the 
Minimum Description Length (MDL) principle. 
MDL \cite{Rissanen89} is a criterion for data compression and statistical 
estimation proposed in information theory.

In applying MDL, we calculate the code length
for encoding each model, referred to as the `model description length'
$L(M)$, the code length for encoding the given data through the model,
referred to as the `data description length' $L({\cal S}|M)$ and their
sum: \[ L(M,{\cal S}) = L(M) + L({\cal S}|M). \] The MDL principle
stipulates that, for both data compression and statistical estimation,
the best probability model with respect to {\em given data} is that
which requires the least total description length.

The data description length is calculated as 
\[
L({\cal S}|M) = - \sum_{(n,v) \in {\cal S}} \log \hat{P}(n,v),
\]
where $\hat{P}$ stands for the maximum likelihood estimate of
$P$ (as defined in Section~\ref{sec:estimate}). 

We then calculate the model description length as 
\[
  L(M) = \frac{k}{2} \log m, \] where $k$ denotes the
number of {\em free} parameters in the model, and $m$ the entire data
size.\footnote{We note that there are alternative ways of calculating
the parameter description length. For example, we can separately
encode the different types of probability parameters; the joint
probabilities $P(C_n, C_v)$, and the conditional probabilities
$P(n|C_n)$ and $P(v|C_v)$. Since these alternatives are approximations
of one another asymptotically, here we use only the simplest
formulation. In the full paper, we plan to compare the empirical
behavior of the alternatives.} In this paper, we ignore the code
length for encoding a `discrete model,' assuming implicitly that they
are equal for all models and consider only the description length for
encoding the parameters of a model as the model description length.

If computation time were of no concern, we could in principle 
calculate the total description length for each 
model and select the optimal model in terms of MDL. 
Since the number of hard clustering models is of order $O(N^N \cdot V^V)$, 
where $N$ and $V$ denote the size of the noun set and the verb set,
respectively, it would be infeasible to do so. 
We therefore need to devise an efficient algorithm that 
heuristically performs this task. 

\section{Clustering Algorithm}
The proposed algorithm, which we call `2D-Clustering,' iteratively
selects a suboptimal MDL-model from among those hard clustering models
which can be obtained from the current model by merging a noun (or
verb) class pair. As it turns out, the minimum description length
criterion can be reformalized in terms of (average) mutual 
information, and a greedy heuristic algorithm can be 
formulated to calculate, in each 
iteration, the reduction of mutual information which would result from
merging any noun (or verb) class pair, and perform the merge having
the least mutual information reduction, {\em provided that the
  reduction is below a variable threshold}.

\vspace{12pt}
\noindent{2D-Clustering$({\cal S},b_n,b_v)$}\\ 
(${\cal S}$ is the input co-occurrence data, and 
$b_n$ and $b_v$ are positive integers.)
\begin{enumerate}
\item Initialize the set of noun classes $T_n$ and the set of
  verb classes $T_v$ as:
\[
T_n = \{ \{ n \} | n \in {\cal N} \}, T_v = \{ \{ v \} | v \in {\cal V} \},
\]
where 
${\cal N}$ and ${\cal V}$ denote the noun set and the verb set, respectively.
\item Repeat the following three steps: 
\begin{enumerate}
\item execute Merge$({\cal S}, T_n, T_v, b_n)$ to update $T_n$,
\item execute Merge$({\cal S}, T_v, T_n, b_v)$ to update $T_v$,
\item if $T_n$ and $T_v$ are unchanged, go to Step 3.
\end{enumerate}
\item Construct and output a thesaurus for nouns based on the history of $T_n$,
and one for verbs based on the history of $T_v$.
\end{enumerate}

Next, we describe the procedure of `Merge,' as it is being applied to 
the set of noun classes with the set of verb classes fixed. 

\vspace{12pt} 
\noindent{Merge$({\cal S}, T_n, T_v, b_n)$} 
\begin{enumerate} \item For each class pair in $T_n$, calculate the
reduction of mutual information which would result from merging
them. (The details will follow.)  Discard those class pairs whose
mutual information reduction (\ref{eq:implement}) is not less than the
threshold of
\[
\frac{(k_B - k_A)\cdot\log m}{2 \cdot m},
\]
where $m$ denotes the total data size, $k_B$ the number of free parameters
in the model before the merge, and $k_A$ the number of free parameters
in the model after the merge. Sort the remaining class pairs in
ascending order with respect to mutual information reduction.
\item Merge the first $b_n$ class pairs in the 
sorted list.
\item Output current $T_n$.
\end{enumerate}

We perform (maximum of) $b_n$  merges at step 2 for improving efficiency, 
which will result in outputting an at-most $b_n$-ary tree. 
Note that, strictly speaking, once we perform one merge, the model will
change and there will no longer be a guarantee that the remaining merges
still remain justifiable from the viewpoint of MDL.

Next, we explain why the criterion in terms of description length can
be reformalized in terms of mutual information. We denote the 
model before a merge as $M_B$ and the model after the merge as $M_A$. 
According to MDL, $M_A$ should have the least increase 
in data description length
\[
\delta L_{dat} = L({\cal S}|M_A) - L({\cal S}|M_B) > 0,
\]
and at the same time satisfies
\[
\delta L_{dat} < \frac{(k_B-k_A) \log m}{2}.
\]
This is due to the fact that the decrease in model description length
equals
\[ L(M_B) - L(M_A) = \frac{(k_B-k_A) \log m}{2} > 0,
\]
and is identical
for each merge.

In addition, suppose that $M_A$ is obtained by merging two noun
classes $C_i$ and $C_j$ in $M_B$ to a single noun class $C_{ij}$. We in
fact need only calculate the difference between description lengths
with respect to these classes, i.e.,
\[ \begin{array}{lll}
  \delta L_{dat} & = & - \sum_{C_v \in T_v}\sum_{n \in C_{ij},v \in C_v}\log \hat{P}(n,v) \\ 
& & + \sum_{C_v \in T_v}\sum_{n \in C_i,v \in C_v}\log \hat{P}(n,v) \\ 
& & + \sum_{C_v \in T_v}\sum_{n \in C_j, v \in C_v}\log \hat{P}(n,v). \\ 
\end{array}
\]
Now using the identity 
\[
\begin{array}{lll}
P(n,v) & = & \frac{P(n)}{P(C_n)}\cdot \frac{P(v)}{P(C_v)}\cdot P(C_n,C_v) \\
 & = & \frac{P(C_n,C_v)}{P(C_n)\cdot P(C_v)} \cdot P(n)\cdot P(v)
\end{array}
\]
we can rewrite the above as 
\[
\begin{array}{l}
  \delta L_{dat} = -
  \sum_{C_v \in T_v} f(C_{ij},C_v)\log\frac{\hat{P}(C_{ij},C_v)}{\hat{P}(C_{ij})\cdot\hat{P}(C_v)}
  \\ 
  + \sum_{C_v \in T_v}
f(C_i,C_v)\log\frac{\hat{P}(C_i,C_v)}{\hat{P}(C_i)\cdot\hat{P}(C_v)}
  \\ + \sum_{C_v \in T_v}
  f(C_j,C_v)\log\frac{\hat{P}(C_j,C_v)}{\hat{P}(C_j)\cdot\hat{P}(C_v)}. \\ 
\end{array}
\]
Thus, the quantity $\delta L_{dat}$ is equivalent to 
the mutual information reduction times 
the data size.\footnote{Average mutual information between 
$T_n$ and $T_v$ is defined as \[I(T_n,T_v) = \sum_{C_n \in T_n}
    \sum_{C_v \in T_v} \left ( P(C_n,C_v) \log
    \frac{P(C_n,C_v)}{P(C_n)\cdot P(C_v)} \right ). \]
}
We conclude therefore that in our present context, a clustering with
the least data description length increase is equivalent to that with
the least mutual information decrease.

Canceling out $\hat{P}(C_v)$ and replacing the probabilities with their maximum likelihood estimates, we obtain 
\begin{equation}\label{eq:implement}
 \begin{array}{l}
  \frac{1}{m} \delta L_{dat} = \frac{1}{m} \biggl( 
  - \sum_{C_v \in T_v} (f(C_i,C_v)+f(C_j,C_v)) \\
\log\frac{f(C_i,C_v)+f(C_j,C_v)}{f(C_i)+f(C_j)} \\ 
  + \sum_{C_v \in T_v}f(C_i,C_v)\log\frac{f(C_i,C_v)}{f(C_i)} \\ 
  + \sum_{C_v \in T_v} f(C_j,C_v)\log\frac{f(C_j,C_v)}{f(C_j)} \biggr). \\ 
\end{array}
\end{equation}
Therefore, we need calculate only this quantity for each possible merge 
at Step 1 of Merge. 

In our implementation of the algorithm, we first load the co-occurrence
data into a matrix, with nouns corresponding to rows, verbs to
columns. When merging a noun class in row $i$ and that in row $j$
($i < j$), for each $C_v$ we add $f(C_i,C_v)$ and $f(C_j,C_v)$
obtaining $f(C_{ij},C_v)$, write $f(C_{ij},C_v)$ on row $i$, move
$f(C_{last},C_v)$ to row $j$, and reduce the matrix by one row.

By the above implementation, the worst case time complexity of the
algorithm is $O(N^3 \cdot V + V^3 \cdot N)$ where $N$
denotes the size of the noun set, $V$ that of the verb set. If we can
merge $b_n$ and $b_v$ classes at each step, the algorithm will become
slightly more efficient with the time complexity of $O(\frac{N^3}{b_n}
\cdot V + \frac{V^3}{b_v} \cdot N)$.

\section{Related Work} \subsection{Models} We can restrict the hard
clustering model (\ref{eq:hardmodel}) by assuming that words within a
same class are generated with an equal probability, obtaining \[
P(n,v) = P(C_n,C_v) \cdot \frac{1}{|C_n|} \cdot \frac{1}{|C_v|},\] which is equivalent to the model proposed
by \cite{Li96b}. Employing this restricted model has the undesirable
tendency to classify into different classes those words that have
similar co-occurrence patterns but have different absolute
frequencies.

The hard clustering model defined in (\ref{eq:hardmodel}) can also be
considered to be an extension of the model proposed by Brown et al. First,
dividing (\ref{eq:hardmodel}) by $P(v)$, we obtain 
\begin{equation}\label{eq:hardmodel2} 
\begin{array}{c}
\frac{P(n,v)}{P(v)} =
P(C_n|C_v) \cdot P(n|C_n) \cdot \left(\frac{P(C_v) \cdot
P(v|C_v)}{P(v)}\right), \\ 
\end{array}
\end{equation} 
Since hard clustering implies $\frac{P(C_v) \cdot P(v|C_v)}{P(v)}=1$
holds, we have \[ P(n|v) = P(C_n|C_v) \cdot P(n|C_n). \] In this way, the hard clustering model turns out to be a
class-based bigram model and is similar to Brown et al's model. The
difference is that the model of (\ref{eq:hardmodel2}) assumes that the
clustering for $C_n$ and the clustering for $C_v$ can be different,
while the model of Brown et al assumes that they are the same.

A very general model of noun verb joint probabilities
is a model of the following form:
\begin{equation}\label{eq:softmodel}
P(n,v) = \sum_{C_n \in \Gamma_n} \sum_{C_v \in \Gamma_v} P(C_n,C_v)\cdot
P(n|C_n) \cdot P(v|C_v).
\end{equation}
Here $\Gamma_n$ denotes a set of noun classes satisfying $\cup_{C_n
\in \Gamma_n} C_n = {\cal N}$, but not necessarily disjoint. 
Similarly $\Gamma_v$ is a set of not necessarily disjoint verb classes.
We can view the problem of clustering words in general as estimation
of such a model. This type of clustering in which 
a word can belong to several different classes
is generally referred to as `soft clustering.' 
If we assume in the above model that each verb forms a verb class by
itself, then (\ref{eq:softmodel}) becomes
\[
P(n,v) = \sum_{C_n \in \Gamma_n} P(C_n,v) \cdot P(n|C_n),
\]
which is equivalent to the model of Pereira et al. 
On the other hand, if we restrict the general model of 
(\ref{eq:softmodel}) so that both noun classes and verb classes
are disjoint, then we obtain the hard clustering model we propose 
here (\ref{eq:hardmodel}).  All of these models, therefore, are 
some special cases of (\ref{eq:softmodel}). 
Each specialization comes with its merit and demerit.  
For example, 
employing a model of soft clustering will make the clustering process 
more flexible but also make the learning process more 
computationally demanding.  Our choice of hard clustering obviously 
has the merits and demerits of the soft clustering model reversed.  

\subsection{Estimation criteria}
Our method is also an extension of that proposed by Brown et al from
the viewpoint of estimation criterion.  Their method 
merges word classes so that the reduction in mutual
information, or equivalently the increase 
in data description length,  is minimized. 
Their method has the tendency to overfit the training data, 
since it is based on MLE. Employing MDL can help solve this problem. 

\section{Disambiguation Method}\label{sec:disam}
We apply the acquired word classes, or more specifically the 
probability model of co-occurrence, to the problem of structural 
disambiguation. In particular, we consider the
problem of resolving pp-attachment ambiguities in quadruples, like
(see, girl, with, telescope) and that of resolving ambiguities in
compound noun triples, like (data, base, system). In the former, we 
determine to which of `see' or `girl' the phrase `with telescope' should be
attached.  In the latter, we judge to which of `base' or `system' the
word `data' should be attached.

We can perform pp-attachment disambiguation by comparing the probabilities
\begin{equation}\label{eq:probs}
\hat{P}_{\rm with}({\rm telescope} | {\rm see}), \hat{P}_{\rm
with}({\rm telescope} | {\rm girl}).
\end{equation}
If the former is larger, we attach `with telescope' to `see;' if the
latter is larger we attach it to `girl;' otherwise we make no decision.
(Disambiguation on compound noun triples can be performed similarly.)

Since the number of probabilities to be estimated is extremely large,
estimating all of these probabilities accurately is generally
infeasible (i.e., the data sparseness problem). Using our clustering
model to calculate these conditional probabilities (by normalizing the
joint probabilities with marginal probabilities) can solve this
problem.

We further enhance our disambiguation method by the following back-off
procedure: We first estimate the two probabilities in question using hard
clustering models constructed by our method. We also estimate the
probabilities using an existing (hand-made) thesaurus with the `tree
cut' estimation method of \cite{Li95}, and use these probability
values when the probabilities estimated based on hard clustering
models are both zero. Finally, if both of them are still zero, we make
a default decision.

\section{Experimental Results}
\subsection{Qualitative evaluation}
In this experiment, we used heuristic rules to extract verbs and
the head words of their direct objects from the {\em tagged} texts of
the WSJ corpus (ACL/DCI CD-ROM1) consisting of 126,084 sentences.

\begin{figure}[htb]
\begin{center}
\epsfxsize6cm\epsfysize6cm\epsfbox{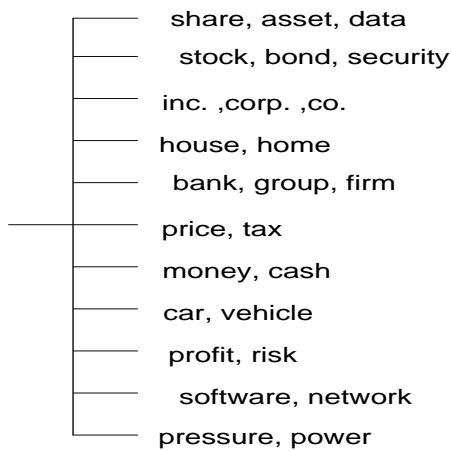}
\vspace{-5pt}
\caption{A part of a constructed thesaurus}
\label{fig:exthesau}
\end{center}
\end{figure}

We then constructed a number of thesauruses based on these data, using
our method. Fig.~\ref{fig:exthesau} shows a part of a thesaurus for
100 randomly selected nouns, based on their appearances as direct 
objects of 20  randomly selected verbs. The thesaurus seems to
agree with human intuition to some degree, although it is constructed
based on a relatively small amount of co-occurrence data. For example,
`stock,' `security,' and `bond' are classified together, despite the fact
that their absolute frequencies in the data vary a great deal
(272, 59, and 79, respectively.) 
The results demonstrate a desirable feature of our
method, namely, it classifies words based solely on the similarities
in co-occurrence data, and is not affected by the absolute
frequencies of the words.

\subsection{Compound noun disambiguation} We extracted compound noun
doubles (e.g., `data base') from the tagged texts of the WSJ
corpus and used them as training data, and then conducted 
structural disambiguation on compound noun triples 
(e.g., `data base system'). 

We first randomly selected 1,000 nouns from the corpus, and extracted
compound noun doubles containing those nouns as 
training data and compound noun triples containing those nouns as test
data. There were 8,604 training data and 299 test data. We hand-labeled
the test data with the correct disambiguation `answers.'

We performed clustering on the nouns on the left position and the
nouns on the right position in the training data by using both 
our method (`2D-Clustering') and Brown et al's method (`Brown'). 
We actually implemented an extended version of their method, which
separately conducts clustering for nouns on the left and those on
the right (which should only improve the performance).

\begin{figure}[htb] 
\begin{center}
\epsfxsize7cm\epsfysize4.5cm\epsfbox{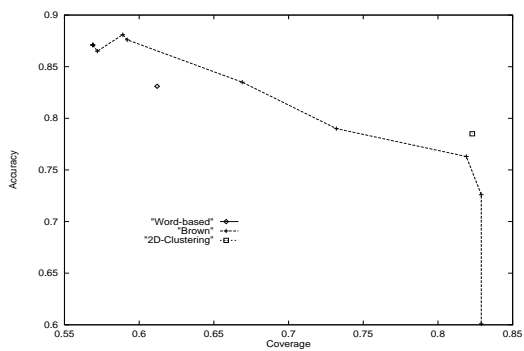}
\vspace{-5pt}
\caption{Compound noun disambiguation results} 
\label{fig:compound} 
\end{center}
\end{figure}

We next conducted structural disambiguation on the test data, using
the probabilities estimated based on 2D-Clustering and Brown. We also
tested the method of using the probabilities estimated based on word
co-occurrences, denoted as `Word-based.' Fig.~\ref{fig:compound} shows
the results in terms of accuracy and coverage, where coverage refers
to the percentage of test data for which the disambiguation method was
able to make a decision. Since for Brown the number of classes finally
created has to be designed in advance, we tried a number of
alternatives and obtained results for each of them.  (Note that, for
2D-Clustering, the optimal number of classes is automatically selected.)

\begin{table}[htb]
\caption{Compound noun disambiguation results}
\vspace{-5pt}
\label{tab:compound}
\begin{center}
\begin{tabular}{|lc|} \hline
Method & Acc.($\%$) \\ \hline
Default & $59.2$ \\
Word-based + Default & $73.9$ \\
Brown + Default & $77.3$ \\
2D-Clustering + Default & $78.3$ \\ \hline
\end{tabular}
\end{center}
\end{table}

Tab.~\ref{tab:compound} shows the final results of all of the above methods
combined with `Default,' in which we attach the first noun to the
neighboring noun when a decision cannot be made by each of the
methods. We see that 2D-Clustering+Default performs the best. These
results demonstrate a desirable aspect of 2D-Clustering, namely, its
ability of {\em automatically} selecting the most appropriate level of
clustering, resulting in neither over-generalization nor
under-generalization.

\subsection{PP-attachment disambiguation} We extracted triples (e.g.,
`see, with, telescope') from the bracketed data of the WSJ corpus
(Penn Tree Bank), and conducted PP-attachment disambiguation on
quadruples. We randomly generated ten sets of data consisting of different
training and test data and conducted experiments through `ten-fold
cross validation,' i.e., all of the experimental results reported
below were obtained by taking average over ten trials.

\begin{table}[htb]
\caption{PP-attachment disambiguation results}
\vspace{-5pt}
\label{tab:pp-attach}
\begin{center}
\begin{tabular}{|lcc|} \hline
Method & Cov.($\%$) & Acc.($\%$) \\ \hline
Default & $100$ & $56.2$ \\
Word-based & $32.3$ & $95.6$ \\
Brown & $51.3$ & $98.3$ \\
2D-Clustering & $51.3$ & $98.3$ \\
Li-Abe96 & $37.3$ & $94.7$ \\
WordNet & $74.3$ & $94.5$ \\
NounClass-2DC & $42.6$ & $97.1$ \\ 
\hline
\end{tabular}
\end{center}
\end{table}

We constructed word classes using our method (`2D-Clustering') and the
method of Brown et al (`Brown').  For both methods, following the
proposal due to \cite{Tokunaga95}, we separately conducted
clustering with respect to each of the 10 most frequently occurring
prepositions (e.g., `for,' `with,' etc). We did not cluster words for
rarely occurring prepositions. We then performed disambiguation based
on 2D-Clustering and Brown. We also tested the method of using the
probabilities estimated based on word co-occurrences, denoted as
`Word-based.'

Next, rather than using the conditional probabilities estimated by our
method, we only used the noun thesauruses constructed by our method,
and applied the method of \cite{Li95} to estimate the best `tree cut
models' within the thesauruses\footnote{The method of \cite{Li95}
outputs a `tree cut model' in a given thesaurus with conditional
probabilities attached to all the nodes in the tree cut. They use MDL
to select the best tree cut model.}  in order to estimate the
conditional probabilities like those in (\ref{eq:probs}). We call the
disambiguation method using these probability values `NounClass-2DC.'  We
also tried the analogous method using thesauruses constructed by 
the method of \cite{Li96b} and estimating the best tree cut models 
(this is exactly the disambiguation method proposed in that paper).  
Finally, we tried using a hand-made thesaurus, WordNet (this is the 
same as the disambiguation method used in \cite{Li95}). 
We denote these methods as `Li-Abe96' and `WordNet,' respectively.

Tab.~\ref{tab:pp-attach} shows the results for all these methods 
in terms of coverage and accuracy.

\begin{table}[htb]
\caption{PP-attachment disambiguation results}
\vspace{-5pt}
\label{tab:pp-attach-final}
\begin{center}
\begin{tabular}{|lc|} \hline
Method & Acc.($\%$) \\ \hline
Word-based + Default & $69.5$ \\
Brown + Default & $76.2$ \\
2D-Clustering + Default & $76.2$ \\
Li-Abe96 + Default & $71.0$ \\
WordNet + Default & $82.2$ \\  
NounClass-2DC + Default & $73.8$ \\ \hline \hline
2D-Clustering + WordNet + Default & $85.2$ \\
Brill-Resnik & $82.4$ \\ 
\hline
\end{tabular}
\end{center}
\end{table}

We then enhanced each of these methods by using a default rule when a
decision cannot be made, which is indicated as `+Default.' 
Tab.~\ref{tab:pp-attach-final} shows the results of these experiments.

We can make a number of observations from these results. (1)
2D-Clustering achieves a broader coverage than NounClass-2DC. This is
because in order to estimate the probabilities for disambiguation, the
former exploits more information than the latter. (2) For Brown, we
show here only its best result, which happens to be the same as the
result for 2D-Clustering, but in order to obtain this result we had to
take the trouble of conducting a number of tests to find the best
level of clustering. For 2D-Clustering, this was done once and
automatically. Compared with Li-Abe96, 2D-Clustering clearly performs
better. Therefore we conclude that our method improves these previous
clustering methods in one way or another. (3) 2D-Clustering outperforms
WordNet in term of accuracy, but not in terms of coverage.  This seems
reasonable, since an automatically constructed thesaurus is more
domain dependent and therefore captures the domain dependent features
better, and thus can help achieve higher accuracy.  On the other hand,
with the relatively small size of training data we had available, its
coverage is smaller than that of a general purpose hand made
thesaurus. The result indicates that it makes sense to combine
automatically constructed thesauruses and a hand-made thesaurus, as we
have proposed in Section~\ref{sec:disam}.

This method of combining both types of thesauruses
`2D-Clustering+WordNet+Default' was then tested. We see that this method
performs the best. (See Tab.~\ref{tab:pp-attach-final}.) 
Finally, for comparison, we tested the `transformation-based
error-driven learning' proposed in \cite{Brill94}, 
which is a state-of-the-art method for pp-attachment
disambiguation. Tab.~\ref{tab:pp-attach-final} shows the result 
for this method as `Brill-Resnik.' 
We see that our disambiguation method also performs better 
than Brill-Resnik. (Note further that for Brill \& Resnik's method, 
we need to use quadruples as training data, whereas ours only 
requires triples.)

\section{Conclusions} We have proposed a new method of clustering
words based on co-occurrence data. Our method employs a probability
model which naturally represents co-occurrence patterns over word
pairs, and makes use of an efficient estimation algorithm based on the
MDL principle. Our clustering method improves upon the previous
methods proposed by Brown et al and \cite{Li96b}, and furthermore it can
be used to derive a disambiguation method with overall disambiguation
accuracy of $85.2\%$, which improves the performance of a
state-of-the-art disambiguation method.

The proposed algorithm, 2D-Clustering, can be used in practice, as
long as the data size is at the level of the current Penn Tree Bank.
Yet it is still relatively computationally demanding, and thus an
important future task is to further improve on its computational
efficiency.

\section*{Acknowledgement}
We are grateful to Dr.\ S.\ Doi of NEC C\&C Media Res.\ Labs.\ for his
encouragement. We thank Ms.\ Y.\ Yamaguchi of NIS for her programming
efforts.


\begin{thebibliography}{}

\bibitem[\protect\citename{Brill and Resnik}1994]{Brill94}
E. Brill and P. Resnik.
\newblock A rule-based approach to prepositional phrase attachment
  disambiguation.
\newblock {\em Proc. of COLING'94}, pp. 1198--1204.

\bibitem[\protect\citename{Brown \bgroup et al.\egroup }1992]{Brown92}
P. F.~Brown, V. J.~Della~Pietra, P. V.~deSouza, J. C.~Lai, and
  R. L.~Mercer.
\newblock 1992.
\newblock Class-based n-gram models of natural language.
\newblock {\em Comp. Ling.}, 18(4):283--298.

\bibitem[\protect\citename{Li and Abe}1995]{Li95}
H. Li and N. Abe.
\newblock 1995.
\newblock Generalizing case frames using a thesaurus and the {MDL} principle.
\newblock {\em Comp. Ling.}, (to appear).

\bibitem[\protect\citename{Li and Abe}1996]{Li96b}
H. Li and N. Abe.
\newblock 1996.
\newblock Clustering words with the {MDL} principle.
\newblock {\em Proc. of COLING'96}, pp. 4--9.

\bibitem[\protect\citename{Pereira \bgroup et al.\egroup }1993]{Pereira93}
F. Pereira, N. Tishby, and L. Lee.
\newblock 1993.
\newblock Distributional clustering of {E}nglish words.
\newblock {\em Proc. of ACL'93}, pp. 183--190.

\bibitem[\protect\citename{Rissanen}1989]{Rissanen89}
J. Rissanen.
\newblock 1989.
\newblock {\em Stochastic Complexity in Statistical Inquiry}.
\newblock World Scientific Publishing Co., Singapore.

\bibitem[\protect\citename{Tokunaga \bgroup et al.\egroup }1995]{Tokunaga95}
T. Tokunaga, M. Iwayama, and H. Tanaka.
\newblock Automatic thesaurus construction based-on grammatical relations.
\newblock {\em Proc. of IJCAI'95}, pp. 1308--1313.

\end{thebibliography}
\end{document}